\documentclass[runningheads]{llncs}
\usepackage{graphicx}
\usepackage{comment}
\usepackage{amsmath,amssymb}
\usepackage{color}

\usepackage[pagebackref=true,breaklinks=true,letterpaper=true,colorlinks,bookmarks=false]{hyperref}
\usepackage{float}
\usepackage{graphicx}
\usepackage{tabularx}
\usepackage{multirow}
\usepackage{rotating}
\usepackage{wrapfig}
\usepackage{lineno}
\usepackage{epsfig}
\usepackage{nicefrac}      
\usepackage{microtype} 
\usepackage{float}
\usepackage{amsmath}

\begin{document}

\pagestyle{headings}
\mainmatter
\def\ECCVSubNumber{719}  

\title{AUTO3D: Novel view synthesis through unsupervisely learned variational viewpoint and global 3D representation} 


\titlerunning{AUTO3D: Unsupervised novel view synthesis}
%

\author{Xiaofeng Liu\inst{1,4\dag*}\orcidID{0000-0002-4514-2016} \and
Tong Che\inst{2\dag}\orcidID{0000-0001-5354-6961} \and
Yiqun Lu\inst{3\dag}\orcidID{0000-0002-4852-292X} \and
Chao Yang\inst{5}\orcidID{0000-0002-6553-7963} \and
Site Li\inst{6}\orcidID{0000-0002-7221-1814} \and
Jane You\inst{7}\orcidID{0000-0002-8181-4836}}

\authorrunning{X. Liu et al.}
%
\institute{HMS, Harvard University, Boston MA 03315, USA \and
MILA, Université de Montréal, Montréal, Canada \and
Nanjing University of Information Science and Technology, Nanjing 210044, China \and
Fanhan Tech. Inc., Suzhou 215128, China. \and
Facebook AI, Boston MA 03315, USA \and
Carnegie Mellon University, Pittsburgh PA 15213, USA \and
Dept. of Computing, The Hong Kong Polytechnic University, Hong Kong\\$\dag$ Contribute Equally. *Corresponding author \email{liuxiaofengcmu@gmail.com}}
 
\maketitle

\begin{abstract}
This paper targets on learning-based novel view synthesis from a single or limited 2D images without the pose supervision. In the viewer-centered coordinates, we construct an end-to-end trainable conditional variational framework to disentangle the unsupervisely learned relative-pose/rotation and implicit global 3D representation (shape, texture and the origin of viewer-centered coordinates, etc.). The global appearance of the 3D object is given by several appearance-describing images taken from any number of viewpoints. Our spatial correlation module extracts a global 3D representation from the appearance-describing images in a permutation invariant manner. Our system can achieve implicitly 3D understanding without explicitly 3D reconstruction. With an unsupervisely learned viewer-centered relative-pose/rotation code, the decoder can hallucinate the novel view continuously by sampling the relative-pose in a prior distribution. In various applications, we demonstrate that our model can achieve comparable or even better results than pose/3D model-supervised learning-based novel view synthesis (NVS) methods with any number of input views.

\end{abstract}

\section{Introduction}

Novel view synthesis (NVS) \cite{xu2019view} aims at generating novel images with arbitrary viewpoints given one or a few description images of an object. NVS has great potential in computer vision, computer graphics and virtual reality.

Current NVS methods can be grouped into two categories, i.e., geometry- and learning-based methods. The geometry-based methods \cite{flynn2016deepstereo,ji2017deep} are usually challenging to estimate the geometric structure in 3D space with single or very limited 2D input \cite{forsyth2002computer,xu2019view}, and need render for appearance mapping. 

On the other hand, with the popularity of deep generative model \cite{goodfellow2014generative}, learning-based solutions directly generate the image in target view, without the explicitly 3D structure and the 2D rendering. As the 3D model estimation and render module are not necessary, it is promising in a wide range of scenarios \cite{xu2019view}. 

The generative adversarial network (GAN) \cite{goodfellow2014generative} can be used for NVS by discretizing the camera views and learn the view-to-view mapping functions between any two pre-defined views \cite{tran2017disentangled,tian2018cr,cao2018load}. Without 3D understanding, these models cannot generalize unseen views effectively, e.g., trained with $10^{\circ}, 20^{\circ}$ and the model is asked to take a $15^{\circ}$ input or generate the viewpoint of $25^{\circ}$ \cite{xu2019view}.


To address this issue, \cite{flynn2016deepstereo,ji2017deep} resort to the extra 3D information e.g., CAD labels, which are usually expensive or inaccessible. \cite{xu2019view} introduces the Cycle GAN \cite{zhu2017unpaired} to extract pose-invariant feature as implicitly 3D representation. However, all of the aforementioned learning-based methods rely on human-labeled camera pose/viewpoint in their training. Getting these viewpoint labels is costly because  the position of camera and object both need to be measured. Besides, the results are usually noisy \cite{Liu_2019_ICCV2}. A more challenging issue of this approach is that it is sometimes difficult to define the origin of pose for unseen, complex new objects.

Actually, previous NVS works adopt the $object$-$centered$ $coordinates$ \cite{shin2018pixels}, where the shape of objects is represented with a canonical view. For example, shown either a front view
or side view of a car, these approaches set the pre-defined frontal view as the origin and synthesize a view in this pose coordinates. Defining canonical poses can simplify some specific scenarios (e.g., face \cite{feng2018joint}), while it is problematic on many real-world tasks. It requires all the 3D objects to be aligned to a canonical pose, which is hard for a novel object that has not been encountered in the training set \cite{Liu_2019_ICCV2}. 

In contrast, $viewer$-$centered$ $coordinates$ \cite{shin2018pixels,zhang2018learning} propose to represent the shape in a coordinate system that aligns with the viewing perspective of input image. We propose that the origin of NVS can be defined as the input view. In this setting, novel objects and poses can be generalized since it is not required to align canonical poses to 3D models. The manipulation code of relative-pose would be the difference between appearance-describing input and target view, rather than an absolute value in object-centered coordinates. 

Besides, for complex objects, a single image is intrinsically ill-posed to describe the entire appearance information of their objects. Recent learning-based NVS works either hallucinate the blurry results \cite{dosovitskiy2016learning} or use CAD model in training \cite{nguyen2018rendernet}. A straightforward solution to improve NVS quality is to collect several images of the same object taken from different viewpoints. Most learning-based works \cite{tran2017disentangled,olszewski2019transformable} directly average the representation of inputs with the help of pose label. While the multiple inputs can be aligned without pose supervision according to the texture in geometry-based methods.

Motivated by the aforementioned insights, we propose \textbf{an} \textbf{u}nsupervised conditional variational autoencoder framework \textbf{to} achieve NVS in learned viewer-centered coordinates (abbreviated as AUTO3D). In this paper, we propose a method to benefit from both learning- and geometry-based methods while ameliorating their drawback. Our method is essentially a learning-based strategy without the need of the explicitly 3d reconstruction and render, and yet still infers 3D knowledge implicitly. It can automatically disentangle the relative-pose/rotation and a global 3D representation to summarize the other factors (e.g., shape, texture, illumination and the origin of viewer-centered coordinates) without any extra supervision of pose, 3D model or geometry priors of symmetry \cite{barron2014shape,kanazawa2018learning}, and synthesize images of continuous viewpoints. 

Our basic idea coincides with human's way of novel view imagination that we can perform virtual rotation of an implicitly 3D world understanding start from the given view in our mentality \cite{shin2018pixels}. We do not need to define frontal view, have input pose label, and extract view-point independent representation as \cite{xu2019view,tran2017disentangled}.

Besides, the disentanglement based on GANs can be unreliable for its unstable training dynamics what is known as mode collapse \cite{goodfellow2016nips,liu2019feature,che2019deep,liu2018data}. {U}nsupervised conditional $\beta$-variational autoencoder (VAE) adopted here for viewer-centered pose encoding offers a much easier and stable training than GANs \cite{goodfellow2016nips}. Although GAN loss can always be added to enrich the generation details \cite{larsen2015autoencoding}. With end-to-end training, our model simultaneously learns to extract 3D information from appearance-describing images, to disentangle latent pose code, and to synthesize target image with a relative-pose code sampled on a prior distribution (e.g., Gaussian). All of these are achieved in a pose-unsupervised manner. 

Our spatial correlation module (SCM) can take multiple images in a permutation invariant manner to generate a global 3D encoding. Based on the non-local mechanism \cite{wang2018non,zhou2017temporal}, we further explore the spatial clues with Gaussian similarity metric and local diffusion-based complementary-aware formulation.

Since these images provide a complete description of the appearance of the object, we name them as ``appearance-describing" images. Our model extracts the implicitly global 3D representation which provides a global overview of the objects from these appearance describing images. The representation is combined with the latent relative-pose code to synthesize the target image with the viewpoint. In our model, no explicit notion of ``canonical pose" is given by the human labeler. Instead, it infers an implicit origin of viewer-centered coordinates from the appearance describing images, which is usually the average pose of these input images in our experiment observations. Besides, the input pose detection in testing is not required When synthesizing the view with a user-defined degree of rotation. Our contributions can be summarized as: 
   
$\bullet$ We propose a novel learning-based NVS system to synthesize new images in arbitrary views without the supervision of pose. AUTO3D is the first attempt at adapting unsupervisely learned viewer-centered coordinates for NVS.

$\bullet$ A unified conditional variational framework is designed to achieve unsupervisely learned viewer-centered relative-pose encoding and global 3D representation (shape, texture, illumination and the origin of viewer-centered coordinates, etc.).

$\bullet$ Our model is general to take any number of images (from one to many) in a permutation-invariant manner. The complementary information is organized with a pose-unsupervised non-local mechanism beyond simply average.

We extensively evaluate our method on both objects and face NVS benchmarks and obtained comparable or even better performance than the pose/3D model-supervised methods. It can be applied to either a single or multiple inputs.

\section{Related Work}

\noindent \textbf{Geometry-based NVS} tries to explicitly model the 3D structure of objects and project it to 2D space \cite{sturm1996factorization,garg2016unsupervised,flynn2016deepstereo,ji2017deep,pontes2018image2mesh}. However, the estimated point clouds are often not dense enough, especially when handling complicated texture \cite{lin2018learning,pontes2018image2mesh}. \cite{garg2016unsupervised,xie2016deep3d} estimated the
depth instead, but they are designed for binocular situations only. \cite{rematas2016novel,kholgade20143d} proposed exemplar-based models that use large-scale collections of 3D models, and the accuracy largely depends on the variation and complexity of 3D models. \cite{henderson2019learning} proposes to reconstruct the 3D model from a single 2D image without pose annotation, but its voxel setting does not consider the appearance. In contrast, our proposed framework is essentially learning-based without the need for explicit 3D reconstruction \cite{szabo2018unsupervised,wu2018learning,insafutdinov2018unsupervised}.

\noindent \textbf{Learning-based NVS} emerges with the development of convolutional neural networks (CNN) \cite{yang2019towards,liu2017adaptive,liu2018adaptive,liu2018ordinal,liu2020severity,liu2020importance,han2020wasserstein,liu2017line,liu2020unimodal}. Early attempts directly map an input image to a paired target image with an encoder-decoder structure \cite{dosovitskiy2015learning}. \cite{zhou2016view} predicts appearance flow instead of synthesizing pixels from scratch. But it is not able to hallucinate the pixels not contained in the appearance-describing view. \cite{park2017transformation} concatenates an additional image completion network, but its 3D annotation for training is not necessary for our setting.

Recently, GAN \cite{goodfellow2016nips,liu2020disentanglement,he2020classification,he2020image2audio} has been utilized to improve the realism of synthesized images \cite{yang2018image,liu2018normalized,liu2019hard}. The generator learns to hallucinate the missing pixels to make the output realistic. Most methods essentially learn an view-to-view translator \cite{isola2017image,zhu2017unpaired,liu2017unsupervised} between any two pre-defined discrete poses. Without taking the 3D knowledge into account, these methods can only synthesize decent results in several views presented in a training set with pose labels. In contrast, our AUTO3D can synthesize novel viewpoints even if they never appear in the training set and no pose label is given. \cite{xu2019view} proposes to extract view-independent features to implicitly infer the 3D structure with pose supervision in the CycleGAN \cite{zhu2017unpaired}. Indeed, all previous mentioned learning-based NVS require either 3D model or pose label in their training \cite{xu2019view,sun2018multi,olszewski2019transformable,tatarchenko2016multi,zhou2016view,chen2019monocular}. Besides, some methods introduce explicit 3D induction bias, e.g., surfel representation \cite{rajeswar2018pix2scene} and rigid-body transformation \cite{nguyen2019hologan}, but do not work on unseen objects in testing. However, based on a unified conditional variational framework, our AUTO3D learns an implicit global 3D representation on the unsupervisely learned viewer-centered coordinates without any 3D shape and pose supervisions, performing well with unseen objects and views.

Multiple-description NVS has also been investigated to provide more information about the object. Most works \cite{tran2017disentangled,liu2018dependency,liu2019dependency} directly average the representation of each appearance-describing input. \cite{sun2018multi} proposes a sophisticate 3D statistic model to integrate different views. Our spatial-aware self-attention can be a simple and efficient learning-based unified solution to tackle this problem.

\noindent \textbf{Self-attention and non-local filtering.} As attention models gain in popularity, \cite{vaswani2017attention} develops a self-attention mechanism for machine translation. A similar idea is inherited in the non-local algorithm \cite{buades2005non}, which is a classical image denoising technique. The interaction networks are also developed for modeling pair-wise interactions \cite{Liu_2019_ICCV}. Moreover, \cite{wang2018non} proposes to bridge self-attention to the more general non-local filtering operations and use it for action recognition in videos. \cite{zhou2017temporal} proposes to learn temporal dependencies between video frames at multiple time scales. However, we argue that it is essentially tailored for unordered image sets. We further incorporating spatial clues with Gaussian similarity matrix, and local diffusion-based complementary-aware formulation.


\section{Methodology}

\begin{figure*}[t]
  \centering
  \includegraphics[width=1\linewidth]{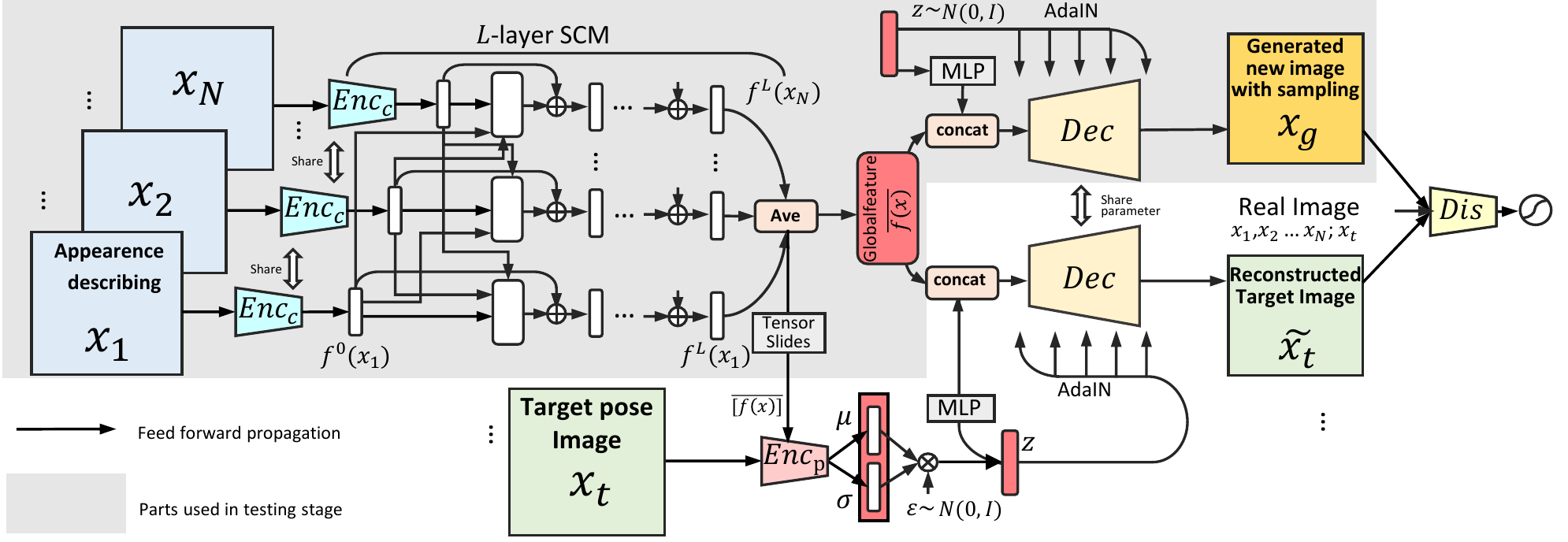} 
  \caption{Illustration of our proposed AUTO3D framework. It is based on VAE-GAN and consists with an unsupervised viewer-centered relative-pose encoding framework, and a spatial-aware self
attention module for global 3D encoding to summarize the other factors. e.g., shape, texture, illumination and the origin of viewer-centered coordinates.}\label{fig2} 
\end{figure*}

Our goal is to generate a novel view image $x_g$ with the controllable viewer-centered relative-pose code $z$ given a global description of the object or scene. The global 3D representation is a vector representation computed from a single or multiple appearance-describing images $\{x_1,x_2\cdots x_N\}, N=1,2\cdots$ which provides a partial or complete view of the 3D object. Our implicit global 3D representation does not pose-invariant as \cite{xu2019view}, since it is used to define the origin of unsupervisely learned viewer-centered coordinates.

The overall framework of our AUTO3D is shown in Fig. \ref{fig2}, which is based on the conditional $\beta$-variational autoencoder. Note that the GAN module is only applied to enrich the details rather than disentanglement. The system is composed of four modules for 1) global 3D feature encoding, 2)unsupervised viewer-centered relative-pose encoding, 3) conditional decoding and 4)discriminating the reconstructed target image with the generated image with $z$ sampling respectively. The disentanglement of relative-viewpoints/rotation and 3D representations can be achieved via the variational framework without the supervision of the 3D model or view-point label, and not relies on adversarial training. Compared with the sophisticate triplet-based adversarial unsupervised disentanglement \cite{mathieu2016disentangling}, our solution is simple but sufficient here.

\subsection{Global 3D encoding with arbitrary number of appearance describing images}

Previous works usually focused on generating 3D model from only a single image \cite{xu2019view}, but it is intrinsically hard to infer the hidden parts from one image for many complex 3D objects. Rather than simply using the average operation to aggregate multiple views \cite{tatarchenko2016multi,zhou2016view,sun2018multi,olszewski2019transformable} without alignment of different views, we propose to use the global 3D encoder to collect the global information of the object. 

The inputs to our global 3D encoder network can be arbitrary number (one to many) of images of the same 3D object taken from different viewpoints, to provide the global information of the 3D object, namely shape, color, texture and the origin of viewer-centered coordinates, etc.

To organize multi-view inputs without the pose label, we first apply the fully convolutional content encoder $Enc_c:x_i\rightarrow\mathbb{R}^{H\times W\times D}$ on each 2D appearance-describing image $x_i$ to extract a compressed representation, where $H$, $W$ and $D$ are the height, width and channel dimension of output feature respectively. In general, the extracted feature is expected to maintain the spatial relationship of each pixel in a 2D image. However, CNN is famous for its spatial invariant property. Following the CoordConv operation \cite{liu2018intriguing}, we concatenate the location of the pixel as two additional channels to the feature map.

Since $Enc_c$ is view-agnostic, simply averaging $Enc_c(x_i)$ does not give a chance for each input to be aware of the others, in order to build links and correspondences between different images, etc. We propose to harvest the spatially-aware inner-set correlations by exploiting the affinity of point-wise feature vectors. We use $i=1,\cdots,{\small H}\times {\small W}$ to index the position in HW plane and the $j$ is the index for all $D$-dimensional feature vectors other than the $i^{th}$ vector ($j=1,\cdots,{\small H}\times {\small W}\times ({\small N}-1))$. Specifically, our non-local block can be formulated as \begin{equation}x_{n\_i}^{l}=x_{n\_i}^{l-1}+\frac{\Omega^l}{C_{n\_i}}\sum_{\forall {n\_j}}\omega(x_{n\_i}^{0},x_{n\_j}^{0})(x_{n\_j}^{l-1}-x_{n\_i}^{l-1})\Delta_{i,j}\nonumber\label{con:2}\end{equation}\begin{align}{C_{n\_i}}=\sum_{\forall n\_j}\omega(x_{n\_i}^{l-1},x_{n\_j}^{l-1})\Delta_{i,j}; l=0,1,\cdots,L \end{align}where ${\Omega^l}\in \mathbb{R}^{1\times1\times D}$ is the weight vector to be learned, $L$ being the number of stacked sub-self attention blocks and ${x}_n^0={x_n}$. The pairwise affinity $\omega(\cdot,\cdot)$ is an scalar. The response is normalized by ${C_{n\_i}}$. The operation of $\omega$ in Eq. \eqref{con:2} is not sensitive to many function choices \cite{wang2018non,zhou2017temporal}. We simply choose the embedded Gaussian given by $\omega(x_{n\_i}^{l-1},x_{n\_j}^{l-1})=e^{\psi{(x_{n\_i}^{l-1})}^T\phi(x_{n\_j}^{l-1})}$, where $\psi({x_{n\_i}^{l-1})={\Psi}x_{n\_i}^{l-1}}$ and $\phi(x_{n\_j}^{l-1})={\Phi}x_{n\_j}^{l-1}$ are two embeddings, and $\Psi$, $\Phi$ are matrices to be learned.

To explore the spatial clues, we further propose to use Gaussian kernel as a similarity measure $\Delta_{i,j}={\rm exp}{(\frac{{\parallel hw_{n\_i}-hw_{n\_j} \parallel}_2^2}{\sigma })}$ , where $hw_{n\_i}$, $hw_{n\_j}\in\mathbb{R}^2$ represent the position of $i^{th}$ and $j^{th}$ vectors in the HW-plane of $x_n$, respectively.

The residual term is the difference between the neighboring feature ($i.e.,x_{n\_j}^{l-1}$) and the computed feature $x_{n\_i}^{l-1}$. If $x_{n\_j}^{l-1}$ incorporates complementary information and has better imaging/content quality compared to $x_{n\_i}^{l-1}$, then RSA will erase some information of the inferior $x_{n\_i}^{l-1}$ and replaces it by the more discriminative feature representation $x_{n\_j}^{l-1}$. Compared to the method of using only $x_{n\_j}^{l-1}$ \cite{wang2018non}, our setting shares more common features with diffusion maps \cite{tao2018nonlocal}, graph Laplacian \cite{chung1997spectral} and non-local image processing \cite{gilboa2007nonlocal}. All of them are non-local analogues \cite{du2012analysis} of local diffusions, which are expected to be more stable than its original non-local counterpart \cite{wang2018non} due to the nature of its inherit Hilbert-Schmidt operator \cite{du2012analysis}.

\subsection{Unsupervised viewer-centered relative-pose encoding}

In the viewer-centered coordinates, the ``average" viewpoint of all the appearance-describing images is defined as origin, while the relative-pose code $z$ indicates the ``rotation" from the origin to the pose of to be synthesized image.

Instead of inferring the viewpoint code only from a target image $x_t$, the viewer-centered relative-pose encoder $Enc_p$ takes both $x_t$ and $\overline{[f(x)]}$ as inputs. $\overline{[f(x)]}$ is a slice of $\overline{f(x)}$. In testing, our latent code $z$ controls how the generated viewpoint is different from the origin w.r.t. a small set of input appearance-describing images.

The $Dec$ maps global 3d feature $\overline{f(x)}$ to image domain with a reversed structure of $Enc$ and conditional to the relative-pose code $z$. Instead of only resize $z$ to match $\overline{f(x)}$ with a multi-layer perceptron (MLP) and concatenate them as the input of $Dec$, we also adopt the adaptive instance normalization (AdaIN) \cite{huang2017arbitrary} after each convolution layer as previous conditional generation works \cite{xu2019view,huang2018introvae,nguyen2019hologan,zakharov2019few}. Specifically, the mean ($\mu$) and variance $\sigma$ of AdaIN layers are normalized to match the relative-pose code $z$ instead of the feature map itself. Here, it a injects stronger inductive bias of $z$ to $Dec$.

The optimization objective of $\beta$-VAE \cite{higgins2017beta} is to maximize the regularized evidence lower bound (ELBO) of $p(x_t|x_1,\cdots x_N)$. Specifically, ${\rm log}p(x_t|x_1,\cdots x_N)\geq E_{q(z|x_t,\overline{[f(x)]})} {\rm log}{p(\tilde{x_t}|z,\overline{f(x)})}-\beta D_{KL}(q(z|x_t,\overline{[f(x)]})||p(z))$, where  $q(z|x_t,\overline{f(x)})$ and $p(\tilde{x_t}|z,\overline{f(x)})$ are the parameterized $Enc$ and $Dec$ respectively, $p(z)$ is a prior distribution (e.g., Gaussian), $D_{KL}$ is the Kullback-Leibler (KL) divergence. The regularization coefficient $\beta\geq1$ constraints the capacity of the latent information bottleneck $z$ \cite{alemi2016deep,saxe2018information}. Therefore, the higher $\beta$ can put a stronger information bottleneck pressure on the latent posterior $q(z|x_t,\overline{[f(x)]})$. In this way, $z$ is forced to contain as little information of $x_t$ as possible, thus it drops all the appearance information and carries only the relative-pose information. Both latent $z$ and $\overline{f(x)}$ are the inputs to the $Dec$. With the information bottleneck on $z$, the decoder is encouraged to get all its appearance information from $\overline{f(x)}$, thus the relative-pose and appearance information are automatically disentangled, without any pose supervision or adversarial training.

We follow the original VAEs \cite{goodfellow2016deep} that the inference model has two output variables, $i.e.,$ $\mu$ and $\sigma$. Then utilize the reparametric trick $z=\mu+\sigma\odot\epsilon$, where $\epsilon\in N(0,I)$. The posterior distribution is $q(z|x_t,\overline{[f(x)]})\sim N(z;\mu,\sigma^2)$. In practice, the KL-divergence can be computed as {\begin{equation}
\begin{aligned} 
    L_{KL}(z;\mu,\sigma)=\frac{1}{2}\sum^{M_z}_{j=1}(1+{\rm log}(\sigma_j^2)-\mu_j^2-\sigma_j^2) \label{eq:m1} 
\end{aligned}\end{equation}}where $M_z$ the dimension of the latent code $z$. For the reconstruction error, we simply adopt the pixel-wise mean square error (MSE), i.e., $L_2$ loss. Let $\tilde{x_t}$ be the reconstructed $x_t$, their $L2$ loss can be formulated as{\begin{equation}
\begin{aligned}
    L_{REC}(x_t,\tilde{x_t})=\frac{1}{2}\sum^{M_{rz}}_{j=1}||x_{t,j}-\tilde{x}_{t,j}||^2_F\label{eq:m2}
\end{aligned}\end{equation}}where ${M_{rz}}$ indicates the channel dimension of $x_t$ or $\tilde{x_t}$.

\subsection{Overall framework and optimization objective}

A limitation of VAEs is that the generated samples tend to be blurry. This is often result of the limited expressiveness of the inference models, the injected noise and imperfect element-wise criteria such as the squared error \cite{larsen2015autoencoding}. Although recent studies \cite{kingma2016improved} have greatly improved the predicted log-likelihood, the VAE image generation quality still lags behind GAN.

In order to improve generation quality, we adopt the following adversarial training procedure. Similar to VAE-GAN \cite{larsen2015autoencoding}, we train AUTO3D to discriminate real samples from both the reconstructions and the generated examples with sampling $z$. As shown in Fig. \ref{fig2}, these two types of samples are the reconstruction
samples $x_r$ and the new samples $\tilde{x_t}$. The adversarial game of GAN can be\begin{equation}
L_{Adv}={\rm log}(Dis({x_r}))+{\rm log}(1-Dis(Dec(x_g)))+{\rm log}(1-Dis(Dec(\tilde{x_t})))
 \label{eq:m1}
\end{equation}where $x_r\in\{x_1,x_2\cdots x_N,x_t\}$ is the real image from either appereance describing set or target pose image. Actually, given a real $x_r$, the reconstructed sample $Dec(\tilde{x_t})$ can always be more realistic than the sampling image $x_g$. We usually use similar number of reconstructed and sampled image in training \cite{larsen2015autoencoding}.

When the KL-divergence object of VAEs is adequately optimized, the posterior $q(z|x_t,\overline{f(x)})$ matches the prior $p(z)= N(z;0,I)$ approximately and the samples are similar to each other. The combined use of samples from $p(z)$ and $q(z|x_t,\overline{[f(x)]})$ is also expected to mitigate the observation gap of $z$ in training and testing stage, and empirically synthesize more realistic samples in the testing. The to be minimized objective of each module are respectively defined as {\begin{equation}
\begin{aligned}
&\mathcal{L}_{Enc_p}=  (L_{Rec}+L_{KL}+L_{Adv}); ~\mathcal{L}_{Dec}= (L_{Rec}+L_{Adv})\\
&\mathcal{L}_{Enc_c/SCM}= (L_{Rec}+L_{Adv}); ~~~~\mathcal{L}_{Dis}=  -L_{Adv} 
\label{eq:m1}
\end{aligned}\end{equation}}After the aforementioned modules are trained, we use $Enc_c$, spatial-aware self attention module (SCM) and $Dec$ for the testing. Give a set of appearance-describing image, we can sampling on a prior $p(z)$ to control the projection view with user defined rotation. Note that the network mapping of $z$ and the relative-pose difference is deterministic after the training. 

\begin{figure}[t]
  \centering
  \includegraphics[width=1\linewidth]{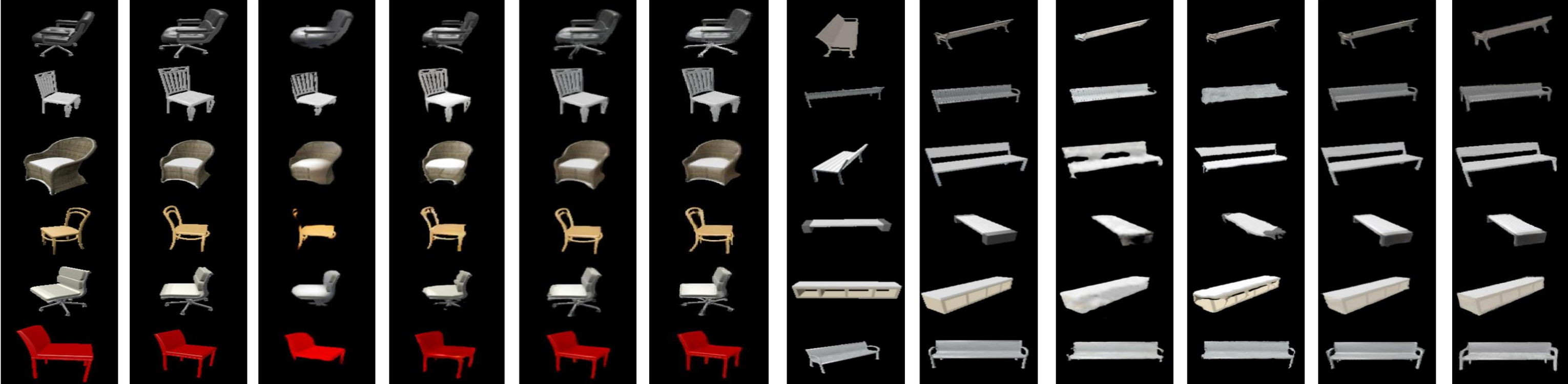} \\
  \caption{Comparison of ``chair, bench" category on ShapeNet with a single 2D input. From left to right: 2D-input, ground-of-truth, MV3D\cite{tatarchenko2016multi}, AF\cite{zhou2016view}, pose-supervised VIGAN\cite{xu2019view}, Our unsupervised AUTO3D. AUTO3D is comparable to the pose-supervised VIGAN and significantly better than MV3D/AF.}\label{fig3}
\end{figure}

\section{Experiments}

We conduct a series of experiments on both large scale objects (ShapeNet \cite{chang2015shapenet}) and face (300W-LP) \cite{zhu2016face} datasets to evaluate the qualitative and quantitative performance of AUTO3D, along with the detailed ablation study. Note that the compared methods use the absolute pose value while our $z$ defines the relative-pose/rotation. For the fair comparison, we calculate the difference of input and target pose label in the testing as our relative-pose. Note that AUTO3D can generate any pose continuously without the pose label in both training and NVS implementation.

For fair comparisons in the objects and continuous face rotation tasks, we choose the same $Enc_c$, $Dec$, $Dis$, MLP backbones and AdaIN setting as VI-GAN \cite{xu2019view}. We set $|z|$=128 for all datasets except for Cars, where we use $|z|$=200. We train AUTO3D from scratch with Adam \cite{kingma2014adam} solver and implemented on Pytorch \cite{paszke2017automatic}. Let $\mathcal{C}_{s,k,c}$ denote a convolutional layer with a stride $s$, kernel size $k$, and an output channel $c$. Then, the discriminator architecture can be expressed as $\mathcal{C}_{2,4,32}\rightarrow\mathcal{C}_{2,4,64}\rightarrow\mathcal{C}_{2,4,128}\rightarrow\mathcal{C}_{2,4,256}\rightarrow\mathcal{C}_{1,1,3}$. Note that we use a local discriminator similar to that of \cite{isola2017image}. We use a Leaky ReLU activation function with slope of 0.2 on every layer, except for the last layer. Normalization layer is not applied. This architecture is shared across all experiments.

We implemented our model on Pytorch \cite{paszke2017automatic}. Our model is trained end-to-end using using ADAM \cite{kingma2014adam} optimization with hyper-parameters $\beta_{1}$=0.9 and $\beta_{2}$=0.999. We used a batch size of 8 for ShapeNet objects. The encoder network is trained using a learning rate of $5\times 10^{-5}$ and the generator is trained using a learning rate $10^{-4}$.

\subsection{Datasets}

ShapeNet \cite{chang2015shapenet} is a large collection of textured 3D CAD models of a variety of object categories. There are both single input setting and multiple inputs setting. For single image only, we use the image rendered by \cite{choy20163d} following \cite{xu2019view}. The chair, bench, and sofa are selected, and 80\% models are used for training while 20\% for testing \cite{xu2019view}. Noticing the testing models are not seen by the network in the training stage. For the multiple viewpoint inputs, we follow the standard training and test data splits 
\cite{tatarchenko2016multi,zhou2016view,park2017transformation,sun2018multi,olszewski2019transformable}, and train a separate network for each object category (also standard), using 1 to 4 input images to synthesize the target view. The network architecture and training methods were fixed across categories.

300W-LP \cite{zhu2016face} is a synthesized large-pose face images from 300W. It generates 61,225 samples across large poses with the 3D Image meshing and rotation of in-the-wild face images, which is further expanded to 122,450 samples with flipping. Following \cite{xu2019view}, we use 80\% identities for training and 20\% for testing.

\subsection{Qualitative results}

Object rotation targets on synthesizing novel views of certain categories for unseen objects. It is challenging, since different objects may have diverse structure and appearance. To demonstrate the capacity of our model, we evaluate our model on the ShapeNet \cite{chang2015shapenet} dataset using samples from “chair”, “bench” and “sofa” categories. The results are given in Fig. \ref{fig3}.

MV3D \cite{tatarchenko2016multi} and Appearance-Flow (AF) \cite{zhou2016view} are two popular methods that perform well on this task, while VI-GAN \cite{xu2019view} is the recent pose-supervised state-of-the-art. MV3D and AF deal with continuous camera pose by taking the difference between the 3$\times$4 transformation matrices of the input and target views as the pose vector. We compare AUTO3D with them both qualitatively and quantitatively. As shown in Figs. \ref{fig3}, MV3D \cite{tatarchenko2016multi} and AF \cite{zhou2016view} usually miss small parts, while our results are closer to the ground truth and recent pose-supervised NVS method.

\begin{table}[t]
\centering
\resizebox{0.7\linewidth}{!}{
\centering
  \begin{tabular}{c|cc|cc|cc}
  \hline
Method&\multicolumn{2}{c|}{Chair}&\multicolumn{2}{c|}{Bench}&\multicolumn{2}{c}{Sofa}\\\cline{2-7}
  
& $L_1\downarrow$ & SSIM$\uparrow$ & $L_1\downarrow$ & SSIM$\uparrow$ & $L_1\downarrow$ & SSIM$\uparrow$ \\\hline\hline

MV3D\cite{tatarchenko2016multi}[need pose label]&24.25 & 0.76 & 20.24 & 0.75  &17.52 & 0.73\\
AF\cite{zhou2016view}[need pose label]&18.44  & 0.82 &  14.42 &  0.85 & 13.26 &  0.77   \\
VIGAN\cite{xu2019view}[need pose label]&\textbf{12.56} & \textbf{0.87} & \textbf{11.52} & \textbf{0.88} &  \textbf{10.13} & \textbf{0.83}  \\\hline\hline
AUTO3D w/o AdaIN&12.65  &0.83 & 11.88  &0.85  & 10.39 & 0.79  \\
AUTO3D w/o GAN &12.64  &0.83 & 11.86  &0.85  & 10.40 & 0.78  \\
AUTO3D w/o TS&12.65  &0.85 & 11.83  &0.86  & 10.35 & 0.80  \\
AUTO3D w/o SCM&\underline{12.62}  & {0.86} & \underline{11.80}  &\underline{0.87} & 10.31 & \underline{0.82} \\

AUTO3D        &\underline{12.62}  &\textbf{0.87} & \underline{11.80}  &\underline{0.87}  & \underline{10.30} & \underline{0.82}  \\
\hline
  \end{tabular}}
  \caption{ Using a single input, the mean pixel-wise $L_1$ error (lower is better) and SSIM (higher is better) between ground truth and predictions generated by previous pose-supervised methods and different AUTO3D settings. When computing the $L_1$ error, pixel values are in range of [0, 255]. The best are bolded, while the second best are underlined.}\label{tab1} 
\end{table}

\begin{figure}[t]
  \centering
  \includegraphics[width=1\linewidth]{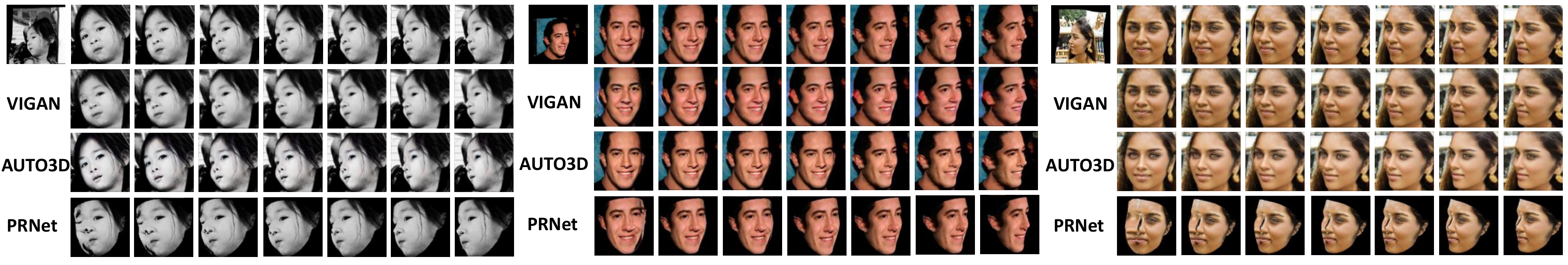} \\
  \caption{Comparison with VIGAN \cite{xu2019view}, PRNet \cite{feng2018joint} on 300W-LP face dataset.}\label{fig7}
\end{figure}

In the face rotation task, PRNet \cite{feng2018joint} uses the UV position map in 3DMM to record 3D coordinates and trains CNN to regress them from single views. Fig. \ref{fig7} qualitatively compares our method with PRNet \cite{feng2018joint} and pose-supervised VI-GAN \cite{xu2019view}. Following \cite{feng2018joint,xu2019view}, we choose the standard training protocol of 300W-LP, but not use the pose label. As shown in Fig. \ref{fig7}, PRNet \cite{feng2018joint} may introduce artifacts when information of certain regions is missing. This issue is severe when turning a profile into a frontal face. In contrast, our model produces more realistic images than PRNet \cite{feng2018joint} and comparable to pose-supervised VI-GAN \cite{xu2019view}.

\subsection{Quantitative results}

For quantitative evaluation, the mean pixel-wise $L_1$ error and the structural similarity index measure (SSIM) \cite{wang2004image,xie2016deep3d} between synthesized results and the ground truth are calculated following previous methods. We measure the capability of our approach to synthesize new views of objects under large transformations following the standard evaluation protocol.

Table \ref{tab1} shows that our model has on-par performance with pose-supervised VI-GAN in single-input setting following their experiment setting. AUTO3D achieves  much lower $L_1$ error and  higher SSIM than MV3D \cite{tatarchenko2016multi} and AF \cite{zhou2016view}.

\begin{table}[t]
\centering
\resizebox{\linewidth}{!}{
\centering
  \begin{tabular}{c|c|cc|cc||c|c|cc|cc}
\hline
Views&Method&\multicolumn{2}{c|}{Chair}&\multicolumn{2}{c||}{Car} & Views&Method&\multicolumn{2}{c|}{Chair}&\multicolumn{2}{c}{Car}    \\\cline{3-6}\cline{8-12}
  
&& $L_1\downarrow$ & SSIM$\uparrow$ & $L_1\downarrow$ & SSIM$\uparrow$& && $L_1\downarrow$ & SSIM$\uparrow$ & $L_1\downarrow$ & SSIM$\uparrow$\\\hline\hline

\multirow{7}*{1}&MV3D\cite{tatarchenko2016multi}[pose]&0.223 & 0.882 & 0.139 & 0.875  &\multirow{7}*{3}&MV3D\cite{tatarchenko2016multi}[pose]&0.197 & 0.898 & 0.116 & 0.887  \\
&AF\cite{zhou2016view}[pose]&0.229  & 0.871 &  0.148 &  0.877  &&AF\cite{zhou2016view}[pose]&0.188  & 0.887 &  0.089 &  0.915    \\

&MNV\cite{sun2018multi}[pose]&0.181  & \textbf{0.895} &  0.098 &  \underline{0.923 } & &MNV\cite{sun2018multi}[pose]&0.122  & 0.919 &  0.068 &  0.941\\
&TBN\cite{olszewski2019transformable}[pose]&\textbf{0.178} & \textbf{0.895} &\textbf{ 0.091} & \textbf{0.927}& &TBN\cite{olszewski2019transformable}[pose]&\textbf{0.116} & \textbf{0.936} & \textbf{0.063} & \textbf{0.943}     \\\cline{2-6}\cline{8-12}

&AUTO3D w/o SCM&\underline{0.181}  &\underline{0.893} & 0.096  &0.916  & &AUTO3D w/o SCM&0.124  & \underline{0.930 }& 0.068  &0.935 \\
&AUTO3D [SCM-SG]& {0.183}  & {0.892} & 0.096  &0.916  & &AUTO3D [SCM-SG]&0.122  & {0.932 }& 0.066  &0.939 \\
&AUTO3D [SCM-LDC]&\underline{0.181}  &\underline{0.893} & 0.096  &0.917  & &AUTO3D [SCM-LDC]&0.120  & \underline{0.934 }& \underline{0.065} &0.939 \\
&AUTO3D     &0.182 &\underline{ 0.893}& \underline{0.095}  &0.916 &&AUTO3D& \underline{0.118}  &\textbf{0.936} & \textbf{0.063}  & \underline{0.942}   \\

\hline\hline

\multirow{7}*{2}&MV3D\cite{tatarchenko2016multi}[pose]&0.209 & 0.890 & 0.124 & 0.883 & \multirow{7}*{4}&MV3D\cite{tatarchenko2016multi}[pose]&0.192 & 0.900 & 0.112 & 0.890 \\
&AF\cite{zhou2016view}[pose]&0.207  & 0.881 &  0.107 &  0.901  & &AF\cite{zhou2016view}[pose]&0.165  & 0.891 &  0.081 &  0.924   \\
&MNV\cite{sun2018multi}[pose]&0.141  & 0.911 &  0.078 &  0.935  &&MNV\cite{sun2018multi}[pose]&0.111  & 0.925 &  0.062 &  \textbf{0.946 }   \\
&TBN\cite{olszewski2019transformable}[pose]&\textbf{0.136} & \textbf{0.928} & \textbf{0.072} & \textbf{0.939} &&TBN\cite{olszewski2019transformable}[pose]& \underline{0.107} &\textbf{ 0.939 }& \textbf{0.059} & \textbf{0.946} \\\cline{2-6}\cline{8-12}

&AUTO3D w/o SCM&0.141  &0.918 & 0.078  &0.929  &&AUTO3D w/o SCM&0.112  &0.929 & 0.062  &0.938  \\

&AUTO3D [SCM-SG]&0.140  &0.921 & 0.076  &0.933 &&AUTO3D [SCM-SG]&0.110  &0.935 & 0.062  &0.942  \\
&AUTO3D [SCM-LDC]&0.139  &0.922 & 0.076  &0.934  &&AUTO3D [SCM-LDC]&0.108  &0.936 & 0.061  &0.944  \\

&AUTO3D&\underline{0.138}  &\underline{0.924} & \underline{0.074}  &\underline{0.937} &&AUTO3D&\textbf{0.106}  & \underline{0.938} &  \underline{0.060 } &\textbf{0.946 }   \\\hline
  \end{tabular}}
  \caption{The mean pixel-wise $L_1$ error (lower is better) and SSIM (higher is better) of AUTO3D and pose-supervised methods with 1 to 4 views, on Chair and Car categories of ShapeNet. Noticing that the setting is different from Table \ref{tab1} as detailed in Sec 4.2.}\label{tab3}
\end{table}

\begin{table}[t]
\centering
\resizebox{1\linewidth}{!}{
\centering
  \begin{tabular}{c|c|c|c}
\hline
Pre-training encoder&PRNet [ECCV2018] \cite{feng2018joint} &VIGAN [ICCV2019] \cite{xu2019view}& Our AUTO3D(unsupervised)\\\hline
$L_1\downarrow$&22.65&\textbf{15.32}& \underline{16.25$\pm$ 0.005 }\\\hline
SSIM$\uparrow$&0.65&\textbf{0.73}& \underline{0.71$\pm$ 0.003} \\
\hline
  \end{tabular}}
  \caption{Turning into frontal face task on 300W-LP dataset.}\label{tab2} 
\end{table}

Then, we demonstrate AUTO3D can infer high-quality views flexibly using limited (1-4) input views at testing. We following the experimental protocol of \cite{sun2018multi,olszewski2019transformable} to use up to 4 input images to infer a target image, which is usually challenging for geometry-based NVS. We report the quantitative results on Table \ref{tab3}, and compare our AUTO3D with other works that can take multiple inputs \cite{tatarchenko2016multi,zhou2016view,sun2018multi,olszewski2019transformable}, as well as those only accepting single inputs \cite{park2017transformation}. AUTO3D is comparable or even better than previous pose-supervised methods, especially when more views available. Besides, the gap between AUTO3D and its SCM-free version is usually larger when views increase.

We also give a quantitative evaluation scheme when turning into frontal faces following \cite{xu2019view}. Given a synthesized frontal image, it is aligned to its ground truth followed by cropping into the facial area. Its ground truth is also cropped with the same operation. $L_1$ error and SSIM are calculated between two facial areas and reported in Table \ref{tab2}. AUTO3D yields higher precision than PRNet \cite{feng2018joint} and is comparable to pose-supervised VIGAN \cite{xu2019view} on the 300W-LP dataset.

\subsection{Ablation study of each module}

Based on conditional $\beta$-VAE, our AdaIN, tensor slides (TS), spatial correlation module (SCM) and adversarial loss (GAN) also contribute to the final results. 

From Table \ref{tab1}, \ref{tab3}, we can see that the SCM does not affect the performance of AUTO3D when only a single input is available. While it is critical to achieve better performance in multiple inputs cases as shown in Table \ref{tab3}. Adding SCM can consistently improve the appearance reconstruction. Besides, SCM without spatial-aware Gaussian (SCM-SG) or local diffusion-based complementary-aware formulation (SCM-LDC)  is consistently inferior to the normal SCM, indicating the effectiveness of our modification on vanilla non-local. 

The adversarial loss is utilized to enrich the details and sharpen the appearance. We do not manage to use it for disentanglement as previous unsupervised adversarial training works \cite{mathieu2016disentangling}.

AdaIN also contributes to disentanglement, and improve the generation quality w.r.t. appearance. Noticing that the NVS is usually not sensitive to the tensor slides, while can speed up the training speed by 1.5 times.

\begin{figure}[t]
  \centering
  \includegraphics[width=0.6\linewidth]{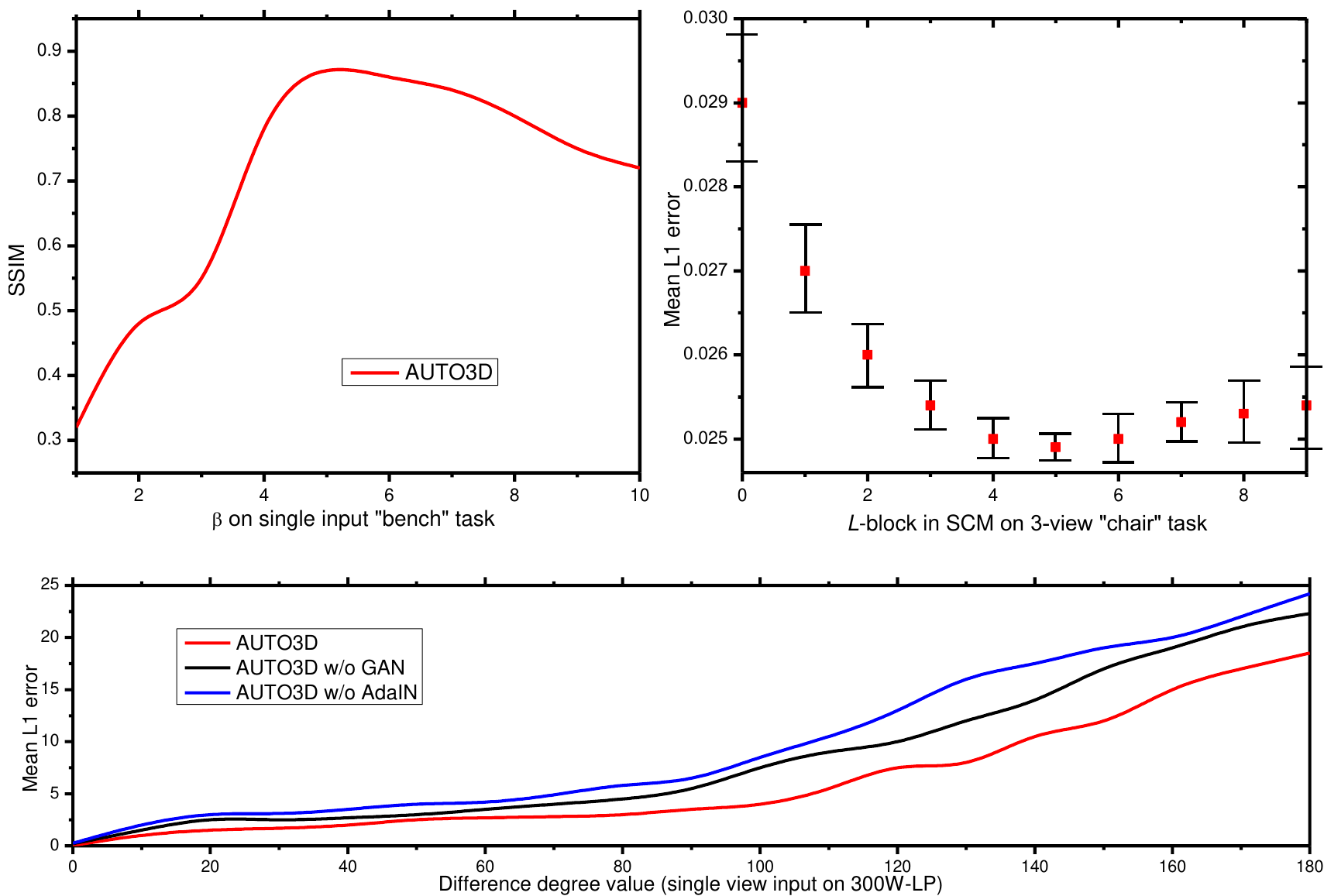} \\
  \caption{Sensibility analysis of [Top left] different $\beta$ (single view bench on ShapeNet), [Top right] the number of SCM blocks (3-view inputs of chair on ShapeNet) and [Bottom] rotation values (single view 300W-LP).}\label{fig8} 
\end{figure}

\subsection{Sensitive analysis}

The value of $\beta$ is critical to the performance. We use automatic selection with the disentanglement metric following \cite{higgins2017beta}, and fine-tune it according to visual quality. The sensitive analysis is shown in top left of Fig. \ref{fig8}.

The number of layers in our spatial correlation module (SCM) is also critical to the synthesis quality in multiple inputs cases. Here, we give a sensibility analysis in the top right of Fig. \ref{fig8} (b). We can see that the performance is stable within the range of $[4,7]$. For simple operation, we choose 4-layer for all of our experiments with multiple inputs.

We also analyse the interaction between the conditional viewer-centered pose code $z$ and generation quality. The bottom of Fig. \ref{fig8} shows the comparisons of $L_1$ error as a function of view rotation on the face dataset. Noticing that $z$ indicates the difference of appearance-describing and target view, and 0$^\circ$ means no viewpoint change. This illustrates that our AUTO3D can well tackle the extreme pose rotations even without the 3D model or pose label in the training.

\subsection{Investigating the global 3D feature} 

We expect that the implicitly 3D structure information of objects can be captured. To evidence this, we implement the experiment of using the latent global 3D representation encoding for learning of 3D tasks. 

Following \cite{xu2019view}, we adopt the 3D face landmark estimation task. The network has two parts where the encoder is the same as the encoder in AUTO3D and Multilayer Perceptron (MLP) is with 2-layers for estimating the coordinate of landmarks based on features extracted by the encoder. Noticing that the backbone of AUTO3D is identical to VIGAN \cite{xu2019view}. We also choose 300W-LP \cite{zhu2016face} for training, in which 3D landmarks are obtained by using their 3DMM parameters.

We configure three training settings to extract the feature for 3D face landmark estimation. The $first$ is to train the overall network from scratch to learn 3D features directly. The $second$ is pre-train the encoder using the view-independent constraint of VI-GAN, then the 3D supervised data is then used to train the overall network. The $third$ setting is to pre-train the $Enc_c$ with our AUTO3D.

Following \cite{xu2019view}, testing involves 2,000 images from AFLW2000-3D \cite{koestinger2011annotated} with 68 landmarks. Besides, the mean Normalized Mean Error (NME) \cite{zhu2016face} is employed for evaluation. We report the results of three settings in Table. \ref{tab4}, until the training loss of both settings no longer changes. The pose-supervised implicitly 3D feature extraction method \cite{xu2019view} and our unsupervised AUTO3D get the mean NMEs of 6.8\% and 6.9\% respectively, which is significantly lower than the training from scratch. This demonstrates that the feature learned by the encoder of AUTO3D is 3D-related. It gives a good initialization for 3D tasks.

\begin{table}[t!]
\centering
\resizebox{0.9\linewidth}{!}{
\centering
  \begin{tabular}{c|c|c|c}
\hline
Pre-train $Enc_c$ &Scratch&VIGAN [ICCV2019] \cite{xu2019view}& Our AUTO3D(unsupervised)\\\hline
mean NMEs$\downarrow$& 12.7\%& \textbf{6.8\%} &  \underline{6.9\%$\pm$0.12\%}\\
\hline
  \end{tabular}} 
  \caption{The NME for 3D face landmark estimation.}\label{tab4} 
\end{table}

\subsection{The Effect of Source Image Ordering}

The sum operation used in AUTO3D is essentially permutation invariant. We conduct a simple experiment where we test the model on all possible order. We randomly sampled 1000 tuple of source (image, camera pose) pairs from ShapeNet cars and chairs, and evaluated on all 24 ordering. We have found that feeding the different order does not affect the performance of proposed AUTO3D. Our model shows robustness to ordering.

\section{Conclusions}

This paper presents a novel learning-based framework (AUTO3D) to achieve NVS without the supervision of pose labels and 3D models. It is essentially based on a conditional $\beta$-VAE which can be easily and stably trained to disentangle the relative viewpoint information from the other factors in global 3D representation (shape, appearance, lighting and the origin of viewer-centered coordinates, etc.). Instead of the conventional object-centered coordinates, we define the relative-pose/rotation in viewer-centered coordinates, for the first time, on NVS task. Therefore, we do not need to align both training exemplars and unseen objects in testing to a pre-defined canonical pose. Both single or multiple inputs can be naturally integrated with a spatial-aware self-attention (SCM) module. Our results evidenced that AUTO3D is a powerful and versatile unsupervised method for NVS. In the future, we plan to explore more 3D tasks with AUTO3D.

\section{Acknowledgements}

This work was supported by the Jangsu Youth Programme [grant number SBK2020041180], National Natural Science Foundation of China, Younth Programme [grant number 61705221], the Fundamental Research Funds for the Central Universities [grant number GK2240260006], NIH [NS061841, NS095986], Fanhan Technology, and Hong Kong Government General Research Fund GRF (Ref. No.152202/14E) are greatly appreciated.

\clearpage
%
%
\bibliographystyle{splncs04}
\bibliography{main}

\begin{thebibliography}{10}
\providecommand{\url}[1]{\texttt{#1}}
\providecommand{\urlprefix}{URL }
\providecommand{\doi}[1]{https://doi.org/#1}

\bibitem{alemi2016deep}
Alemi, A.A., Fischer, I., Dillon, J.V., Murphy, K.: Deep variational
  information bottleneck. arXiv preprint arXiv:1612.00410  (2016)

\bibitem{barron2014shape}
Barron, J.T., Malik, J.: Shape, illumination, and reflectance from shading.
  IEEE transactions on pattern analysis and machine intelligence
  \textbf{37}(8),  1670--1687 (2014)

\bibitem{buades2005non}
Buades, A., Coll, B., Morel, J.M.: A non-local algorithm for image denoising.
  In: Computer Vision and Pattern Recognition, 2005. CVPR 2005. IEEE Computer
  Society Conference on. vol.~2, pp. 60--65. IEEE (2005)

\bibitem{cao2018load}
Cao, J., Hu, Y., Yu, B., He, R., Sun, Z.: Load balanced gans for multi-view
  face image synthesis. arXiv preprint arXiv:1802.07447  (2018)

\bibitem{chang2015shapenet}
Chang, A.X., Funkhouser, T., Guibas, L., Hanrahan, P., Huang, Q., Li, Z.,
  Savarese, S., Savva, M., Song, S., Su, H., et~al.: Shapenet: An
  information-rich 3d model repository. arXiv preprint arXiv:1512.03012  (2015)

\bibitem{che2019deep}
Che, T., Liu, X., Li, S., Ge, Y., Zhang, R., Xiong, C., Bengio, Y.: Deep
  verifier networks: Verification of deep discriminative models with deep
  generative models. arXiv preprint arXiv:1911.07421  (2019)

\bibitem{chen2019monocular}
Chen, X., Song, J., Hilliges, O.: Monocular neural image based rendering with
  continuous view control. In: Proceedings of the IEEE International Conference
  on Computer Vision. pp. 4090--4100 (2019)

\bibitem{choy20163d}
Choy, C.B., Xu, D., Gwak, J., Chen, K., Savarese, S.: 3d-r2n2: A unified
  approach for single and multi-view 3d object reconstruction. In: European
  conference on computer vision. pp. 628--644. Springer (2016)

\bibitem{chung1997spectral}
Chung, F.R., Graham, F.C.: Spectral graph theory. No.~92, American Mathematical
  Soc. (1997)

\bibitem{dosovitskiy2016learning}
Dosovitskiy, A., Springenberg, J.T., Tatarchenko, M., Brox, T.: Learning to
  generate chairs, tables and cars with convolutional networks. IEEE
  transactions on pattern analysis and machine intelligence  \textbf{39}(4),
  692--705 (2016)

\bibitem{dosovitskiy2015learning}
Dosovitskiy, A., Tobias~Springenberg, J., Brox, T.: Learning to generate chairs
  with convolutional neural networks. In: Proceedings of the IEEE Conference on
  Computer Vision and Pattern Recognition. pp. 1538--1546 (2015)

\bibitem{du2012analysis}
Du, Q., Gunzburger, M., Lehoucq, R.B., Zhou, K.: Analysis and approximation of
  nonlocal diffusion problems with volume constraints. SIAM review
  \textbf{54}(4),  667--696 (2012)

\bibitem{feng2018joint}
Feng, Y., Wu, F., Shao, X., Wang, Y., Zhou, X.: Joint 3d face reconstruction
  and dense alignment with position map regression network. In: Proceedings of
  the European Conference on Computer Vision (ECCV). pp. 534--551 (2018)

\bibitem{flynn2016deepstereo}
Flynn, J., Neulander, I., Philbin, J., Snavely, N.: Deepstereo: Learning to
  predict new views from the world's imagery. In: Proceedings of the IEEE
  Conference on Computer Vision and Pattern Recognition. pp. 5515--5524 (2016)

\bibitem{forsyth2002computer}
Forsyth, D.A., Ponce, J.: Computer vision: a modern approach. Prentice Hall
  Professional Technical Reference (2002)

\bibitem{garg2016unsupervised}
Garg, R., BG, V.K., Carneiro, G., Reid, I.: Unsupervised cnn for single view
  depth estimation: Geometry to the rescue. In: European Conference on Computer
  Vision. pp. 740--756. Springer (2016)

\bibitem{gilboa2007nonlocal}
Gilboa, G., Osher, S.: Nonlocal linear image regularization and supervised
  segmentation. Multiscale Modeling \& Simulation  \textbf{6}(2),  595--630
  (2007)

\bibitem{goodfellow2016nips}
Goodfellow, I.: Nips 2016 tutorial: Generative adversarial networks. arXiv
  preprint arXiv:1701.00160  (2016)

\bibitem{goodfellow2016deep}
Goodfellow, I., Bengio, Y., Courville, A.: Deep learning. MIT press (2016)

\bibitem{goodfellow2014generative}
Goodfellow, I., Pouget-Abadie, J., Mirza, M., Xu, B., Warde-Farley, D., Ozair,
  S., Courville, A., Bengio, Y.: Generative adversarial nets. In: Advances in
  neural information processing systems. pp. 2672--2680 (2014)

\bibitem{han2020wasserstein}
Han, Y., Liu, X., Sheng, Z., Ren, Y., Han, X., You, J., Liu, R., Luo, Z.:
  Wasserstein loss-based deep object detection. In: Proceedings of the IEEE/CVF
  Conference on Computer Vision and Pattern Recognition Workshops. pp. 998--999
  (2020)

\bibitem{he2020classification}
He, G., Liu, X., Fan, F., You, J.: Classification-aware semi-supervised domain
  adaptation. In: Proceedings of the IEEE/CVF Conference on Computer Vision and
  Pattern Recognition Workshops. pp. 964--965 (2020)

\bibitem{he2020image2audio}
He, G., Liu, X., Fan, F., You, J.: Image2audio: Facilitating semi-supervised
  audio emotion recognition with facial expression image. In: Proceedings of
  the IEEE/CVF Conference on Computer Vision and Pattern Recognition Workshops.
  pp. 912--913 (2020)

\bibitem{henderson2019learning}
Henderson, P., Ferrari, V.: Learning single-image 3d reconstruction by
  generative modelling of shape, pose and shading. International Journal of
  Computer Vision pp. 1--20 (2019)

\bibitem{higgins2017beta}
Higgins, I., Matthey, L., Pal, A., Burgess, C., Glorot, X., Botvinick, M.,
  Mohamed, S., Lerchner, A.: beta-vae: Learning basic visual concepts with a
  constrained variational framework. ICLR  \textbf{2}(5), ~6 (2017)

\bibitem{huang2018introvae}
Huang, H., He, R., Sun, Z., Tan, T., et~al.: Introvae: Introspective
  variational autoencoders for photographic image synthesis. In: Advances in
  Neural Information Processing Systems. pp. 52--63 (2018)

\bibitem{huang2017arbitrary}
Huang, X., Belongie, S.: Arbitrary style transfer in real-time with adaptive
  instance normalization. In: Proceedings of the IEEE International Conference
  on Computer Vision. pp. 1501--1510 (2017)

\bibitem{insafutdinov2018unsupervised}
Insafutdinov, E., Dosovitskiy, A.: Unsupervised learning of shape and pose with
  differentiable point clouds. In: Advances in neural information processing
  systems. pp. 2802--2812 (2018)

\bibitem{isola2017image}
Isola, P., Zhu, J.Y., Zhou, T., Efros, A.A.: Image-to-image translation with
  conditional adversarial networks. In: Proceedings of the IEEE conference on
  computer vision and pattern recognition. pp. 1125--1134 (2017)

\bibitem{ji2017deep}
Ji, D., Kwon, J., McFarland, M., Savarese, S.: Deep view morphing. In:
  Proceedings of the IEEE Conference on Computer Vision and Pattern
  Recognition. pp. 2155--2163 (2017)

\bibitem{kanazawa2018learning}
Kanazawa, A., Tulsiani, S., Efros, A.A., Malik, J.: Learning category-specific
  mesh reconstruction from image collections. In: Proceedings of the European
  Conference on Computer Vision (ECCV). pp. 371--386 (2018)

\bibitem{kholgade20143d}
Kholgade, N., Simon, T., Efros, A., Sheikh, Y.: 3d object manipulation in a
  single photograph using stock 3d models. ACM Transactions on Graphics (TOG)
  \textbf{33}(4),  1--12 (2014)

\bibitem{kingma2014adam}
Kingma, D.P., Ba, J.: Adam: A method for stochastic optimization. arXiv
  preprint arXiv:1412.6980  (2014)

\bibitem{kingma2016improved}
Kingma, D.P., Salimans, T., Jozefowicz, R., Chen, X., Sutskever, I., Welling,
  M.: Improved variational inference with inverse autoregressive flow. In:
  Advances in neural information processing systems. pp. 4743--4751 (2016)

\bibitem{koestinger2011annotated}
Koestinger, M., Wohlhart, P., Roth, P.M., Bischof, H.: Annotated facial
  landmarks in the wild: A large-scale, real-world database for facial landmark
  localization. In: 2011 IEEE international conference on computer vision
  workshops (ICCV workshops). pp. 2144--2151. IEEE (2011)

\bibitem{larsen2015autoencoding}
Larsen, A.B.L., S{\o}nderby, S.K., Larochelle, H., Winther, O.: Autoencoding
  beyond pixels using a learned similarity metric. ICML  (2016)

\bibitem{lin2018learning}
Lin, C.H., Kong, C., Lucey, S.: Learning efficient point cloud generation for
  dense 3d object reconstruction. In: Thirty-Second AAAI Conference on
  Artificial Intelligence (2018)

\bibitem{liu2017unsupervised}
Liu, M.Y., Breuel, T., Kautz, J.: Unsupervised image-to-image translation
  networks. In: Advances in neural information processing systems. pp. 700--708
  (2017)

\bibitem{liu2018intriguing}
Liu, R., Lehman, J., Molino, P., Such, F.P., Frank, E., Sergeev, A., Yosinski,
  J.: An intriguing failing of convolutional neural networks and the coordconv
  solution. In: Advances in Neural Information Processing Systems. pp.
  9605--9616 (2018)

\bibitem{liu2020disentanglement}
Liu, X.: Disentanglement for discriminative visual recognition. arXiv preprint
  arXiv:2006.07810  (2020)

\bibitem{liu2018dependency}
Liu, X., B.V.K, K., Yang, C., Tang, Q., You, J.: Dependency-aware attention
  control for unconstrained face recognition with image sets. In: European
  Conference on Computer Vision (2018)

\bibitem{liu2020unimodal}
Liu, X., Fan, F., Kong, L., Diao, Z., Xie, W., Lu, J., You, J.: Unimodal
  regularized neuron stick-breaking for ordinal classification. Neurocomputing
  (2020)

\bibitem{liu2018adaptive}
Liu, X., Ge, Y., Yang, C., Jia, P.: Adaptive metric learning with deep neural
  networks for video-based facial expression recognition. Journal of Electronic
  Imaging  \textbf{27}(1),  013022 (2018)

\bibitem{liu2019dependency}
Liu, X., Guo, Z., Jia, J., Kumar, B.: Dependency-aware attention control for
  imageset-based face recognition. In: IEEE Transactions on Information
  Forensics and Security (2019)

\bibitem{Liu_2019_ICCV}
Liu, X., Guo, Z., Li, S., Kong, L., Jia, P., You, J., Kumar, B.V.:
  Permutation-invariant feature restructuring for correlation-aware image
  set-based recognition. In: The IEEE International Conference on Computer
  Vision (ICCV) (October 2019)

\bibitem{liu2020importance}
Liu, X., Han, Y., Bai, S., Ge, Y., Wang, T., Han, X., Li, S., You, J., Lu, J.:
  Importance-aware semantic segmentation in self-driving with discrete
  wasserstein training. In: AAAI. pp. 11629--11636 (2020)

\bibitem{liu2020severity}
Liu, X., Ji, W., You, J., Fakhri, G.E., Woo, J.: Severity-aware semantic
  segmentation with reinforced wasserstein training. In: Proceedings of the
  IEEE/CVF Conference on Computer Vision and Pattern Recognition. pp.
  12566--12575 (2020)

\bibitem{liu2017line}
Liu, X., Kong, L., Diao, Z., Jia, P.: Line-scan system for continuous hand
  authentication. Optical Engineering  \textbf{56}(3),  033106 (2017)

\bibitem{liu2018normalized}
Liu, X., Kumar, B.V., Ge, Y., Yang, C., You, J., Jia, P.: Normalized face image
  generation with perceptron generative adversarial networks. In: 2018 IEEE 4th
  International Conference on Identity, Security, and Behavior Analysis (ISBA).
  pp.~1--8 (2018)

\bibitem{liu2019hard}
Liu, X., Kumar, B.V., Jia, P., You, J.: Hard negative generation for
  identity-disentangled facial expression recognition. Pattern Recognition
  \textbf{88},  1--12 (2019)

\bibitem{liu2019feature}
Liu, X., Li, S., Kong, L., Xie, W., Jia, P., You, J., Kumar, B.: Feature-level
  frankenstein: Eliminating variations for discriminative recognition. In:
  Proceedings of the IEEE Conference on Computer Vision and Pattern
  Recognition. pp. 637--646 (2019)

\bibitem{liu2017adaptive}
Liu, X., Vijaya~Kumar, B., You, J., Jia, P.: Adaptive deep metric learning for
  identity-aware facial expression recognition. In: CVPR Workshops. pp. 20--29
  (2017)

\bibitem{Liu_2019_ICCV2}
Liu, X., Zou, Y., Che, T., Ding, P., Jia, P., You, J., Kumar, B.V.:
  Conservative wasserstein training for pose estimation. In: The IEEE
  International Conference on Computer Vision (ICCV) (October 2019)

\bibitem{liu2018data}
Liu, X., Zou, Y., Kong, L., Diao, Z., Yan, J., Wang, J., Li, S., Jia, P., You,
  J.: Data augmentation via latent space interpolation for image
  classification. In: 24th International Conference on Pattern Recognition
  (ICPR). pp. 728--733 (2018)

\bibitem{liu2018ordinal}
Liu, X., Zou, Y., Song, Y., Yang, C., You, J., K~Vijaya~Kumar, B.: Ordinal
  regression with neuron stick-breaking for medical diagnosis. In: Proceedings
  of the European Conference on Computer Vision (ECCV). pp.~0--0 (2018)

\bibitem{mathieu2016disentangling}
Mathieu, M.F., Zhao, J.J., Zhao, J., Ramesh, A., Sprechmann, P., LeCun, Y.:
  Disentangling factors of variation in deep representation using adversarial
  training. In: Advances in neural information processing systems. pp.
  5040--5048 (2016)

\bibitem{nguyen2019hologan}
Nguyen-Phuoc, T., Li, C., Theis, L., Richardt, C., Yang, Y.L.: Hologan:
  Unsupervised learning of 3d representations from natural images. arXiv
  preprint arXiv:1904.01326  (2019)

\bibitem{nguyen2018rendernet}
Nguyen-Phuoc, T.H., Li, C., Balaban, S., Yang, Y.: Rendernet: A deep
  convolutional network for differentiable rendering from 3d shapes. In:
  Advances in Neural Information Processing Systems. pp. 7891--7901 (2018)

\bibitem{olszewski2019transformable}
Olszewski, K., Tulyakov, S., Woodford, O., Li, H., Luo, L.: Transformable
  bottleneck networks. arXiv preprint arXiv:1904.06458  (2019)

\bibitem{park2017transformation}
Park, E., Yang, J., Yumer, E., Ceylan, D., Berg, A.C.: Transformation-grounded
  image generation network for novel 3d view synthesis. In: Proceedings of the
  ieee conference on computer vision and pattern recognition. pp. 3500--3509
  (2017)

\bibitem{paszke2017automatic}
Paszke, A., Gross, S., Chintala, S., Chanan, G., Yang, E., DeVito, Z., Lin, Z.,
  Desmaison, A., Antiga, L., Lerer, A.: Automatic differentiation in pytorch
  (2017)

\bibitem{pontes2018image2mesh}
Pontes, J.K., Kong, C., Sridharan, S., Lucey, S., Eriksson, A., Fookes, C.:
  Image2mesh: A learning framework for single image 3d reconstruction. In:
  Asian Conference on Computer Vision. pp. 365--381. Springer (2018)

\bibitem{rajeswar2018pix2scene}
Rajeswar, S., Mannan, F., Golemo, F., Vazquez, D., Nowrouzezahrai, D.,
  Courville, A.: Pix2scene: Learning implicit 3d representations from images
  (2018)

\bibitem{rematas2016novel}
Rematas, K., Nguyen, C.H., Ritschel, T., Fritz, M., Tuytelaars, T.: Novel views
  of objects from a single image. IEEE transactions on pattern analysis and
  machine intelligence  \textbf{39}(8),  1576--1590 (2016)

\bibitem{saxe2018information}
Saxe, A.M., Bansal, Y., Dapello, J., Advani, M., Kolchinsky, A., Tracey, B.D.,
  Cox, D.D.: On the information bottleneck theory of deep learning  (2018)

\bibitem{shin2018pixels}
Shin, D., Fowlkes, C.C., Hoiem, D.: Pixels, voxels, and views: A study of shape
  representations for single view 3d object shape prediction. In: Proceedings
  of the IEEE Conference on Computer Vision and Pattern Recognition. pp.
  3061--3069 (2018)

\bibitem{sturm1996factorization}
Sturm, P., Triggs, B.: A factorization based algorithm for multi-image
  projective structure and motion. In: European conference on computer vision.
  pp. 709--720. Springer (1996)

\bibitem{sun2018multi}
Sun, S.H., Huh, M., Liao, Y.H., Zhang, N., Lim, J.J.: Multi-view to novel view:
  Synthesizing novel views with self-learned confidence. In: Proceedings of the
  European Conference on Computer Vision (ECCV). pp. 155--171 (2018)

\bibitem{szabo2018unsupervised}
Szab{\'o}, A., Favaro, P.: Unsupervised 3d shape learning from image
  collections in the wild. arXiv preprint arXiv:1811.10519  (2018)

\bibitem{tao2018nonlocal}
Tao, Y., Sun, Q., Du, Q., Liu, W.: Nonlocal neural networks, nonlocal diffusion
  and nonlocal modeling. arXiv preprint arXiv:1806.00681  (2018)

\bibitem{tatarchenko2016multi}
Tatarchenko, M., Dosovitskiy, A., Brox, T.: Multi-view 3d models from single
  images with a convolutional network. In: European Conference on Computer
  Vision. pp. 322--337. Springer (2016)

\bibitem{tian2018cr}
Tian, Y., Peng, X., Zhao, L., Zhang, S., Metaxas, D.N.: Cr-gan: learning
  complete representations for multi-view generation. arXiv preprint
  arXiv:1806.11191  (2018)

\bibitem{tran2017disentangled}
Tran, L., Yin, X., Liu, X.: Disentangled representation learning gan for
  pose-invariant face recognition. In: CVPR. vol.~3, p.~7 (2017)

\bibitem{vaswani2017attention}
Vaswani, A., Shazeer, N., Parmar, N., Uszkoreit, J., Jones, L., Gomez, A.N.,
  Kaiser, {\L}., Polosukhin, I.: Attention is all you need. In: Advances in
  Neural Information Processing Systems. pp. 5998--6008 (2017)

\bibitem{wang2018non}
Wang, X., Girshick, R., Gupta, A., He, K.: Non-local neural networks. In: The
  IEEE Conference on Computer Vision and Pattern Recognition (CVPR) (2018)

\bibitem{wang2004image}
Wang, Z., Bovik, A.C., Sheikh, H.R., Simoncelli, E.P., et~al.: Image quality
  assessment: from error visibility to structural similarity. IEEE transactions
  on image processing  \textbf{13}(4),  600--612 (2004)

\bibitem{wu2018learning}
Wu, J., Zhang, C., Zhang, X., Zhang, Z., Freeman, W.T., Tenenbaum, J.B.:
  Learning shape priors for single-view 3d completion and reconstruction. In:
  Proceedings of the European Conference on Computer Vision (ECCV). pp.
  646--662 (2018)

\bibitem{xie2016deep3d}
Xie, J., Girshick, R., Farhadi, A.: Deep3d: Fully automatic 2d-to-3d video
  conversion with deep convolutional neural networks. In: European Conference
  on Computer Vision. pp. 842--857. Springer (2016)

\bibitem{xu2019view}
Xu, X., Chen, Y.C., Jia, J.: View independent generative adversarial network
  for novel view synthesis. In: Proceedings of the IEEE International
  Conference on Computer Vision. pp. 7791--7800 (2019)

\bibitem{yang2019towards}
Yang, C., Liu, X., Tang, Q., Kuo, C.C.J.: Towards disentangled representations
  for human retargeting by multi-view learning. arXiv preprint arXiv:1912.06265
   (2019)

\bibitem{yang2018image}
Yang, C., Song, Y., Liu, X., Tang, Q., Kuo, C.C.J.: Image inpainting using
  block-wise procedural training with annealed adversarial counterpart. arXiv
  preprint arXiv:1803.08943  (2018)

\bibitem{zakharov2019few}
Zakharov, E., Shysheya, A., Burkov, E., Lempitsky, V.: Few-shot adversarial
  learning of realistic neural talking head models. arXiv preprint
  arXiv:1905.08233  (2019)

\bibitem{zhang2018learning}
Zhang, X., Zhang, Z., Zhang, C., Tenenbaum, J., Freeman, B., Wu, J.: Learning
  to reconstruct shapes from unseen classes. In: Advances in Neural Information
  Processing Systems. pp. 2257--2268 (2018)

\bibitem{zhou2017temporal}
Zhou, B., Andonian, A., Torralba, A.: Temporal relational reasoning in videos.
  In ECCV  (2018)

\bibitem{zhou2016view}
Zhou, T., Tulsiani, S., Sun, W., Malik, J., Efros, A.A.: View synthesis by
  appearance flow. In: European conference on computer vision. pp. 286--301.
  Springer (2016)

\bibitem{zhu2017unpaired}
Zhu, J.Y., Park, T., Isola, P., Efros, A.A.: Unpaired image-to-image
  translation using cycle-consistent adversarial networks. In: Proceedings of
  the IEEE international conference on computer vision. pp. 2223--2232 (2017)

\bibitem{zhu2016face}
Zhu, X., Lei, Z., Liu, X., Shi, H., Li, S.Z.: Face alignment across large
  poses: A 3d solution. In: Proceedings of the IEEE conference on computer
  vision and pattern recognition. pp. 146--155 (2016)

\end{thebibliography}
\end{document}